\documentclass[iicol, pdflatex,sn-mathphys-num]{sn-jnl}% Math and Physical Sciences Numbered Reference Style
\usepackage{graphicx}%
\usepackage{multirow}%
\usepackage{amsmath,amssymb,amsfonts}%
\usepackage{amsthm}%
\usepackage{mathrsfs}%
\usepackage[title]{appendix}%
\usepackage{xcolor}%
\usepackage{textcomp}%
\usepackage{manyfoot}%
\usepackage{booktabs}%
\usepackage{algorithm}%
\usepackage{algorithmicx}%
\usepackage{algpseudocode}%
\usepackage{listings}%
\theoremstyle{thmstyleone}%
\theoremstyle{thmstyletwo}%

\theoremstyle{thmstylethree}%

\usepackage{array}
\newcolumntype{P}[1]{>{\centering\arraybackslash}p{#1}}
\usepackage{algpseudocode}
\usepackage{arydshln}
\usepackage{pifont}
\newcommand{\cmark}{\ding{51}}
\newcommand{\xmark}{\ding{55}}
\usepackage{orcidlink}

\begin{document}

\title[OOS-DSD: Improving Out-of-stock Detection in Retail Images using Auxiliary Tasks]{OOS-DSD: Improving Out-of-stock Detection in Retail Images using Auxiliary Tasks}


\author*{\fnm{Franko} \sur{Šikić}}\email{franko.sikic@fer.unizg.hr}

\author{\fnm{Sven} \sur{Lončarić}}

\affil{\orgname{University of Zagreb Faculty of Electrical Engineering and Computing}, \orgaddress{\street{Unska 3}, \city{Zagreb}, \postcode{10000}, \country{Croatia}}}

\abstract{
Out-of-stock (OOS) detection is a very important retail verification process that aims to infer the unavailability of products in their designated areas on the shelf.
In this paper, we introduce OOS-DSD, a novel deep learning-based method that advances OOS detection through auxiliary learning. In particular, we extend a well-established YOLOv8 object detection architecture with additional convolutional branches to simultaneously detect OOS, segment products, and estimate scene depth.
While OOS detection and product segmentation branches are trained using ground truth data, the depth estimation branch is trained using pseudo-labeled annotations produced by the state-of-the-art (SOTA) depth estimation model Depth Anything V2. Furthermore, since the aforementioned pseudo-labeled depth estimates display relative depth, we propose an appropriate depth normalization procedure that stabilizes the training process.
The experimental results show that the proposed method surpassed the performance of the SOTA OOS detection methods by 1.8\% of the mean average precision (mAP). In addition, ablation studies confirm the effectiveness of auxiliary learning and the proposed depth normalization procedure, with the former increasing mAP by 3.7\% and the latter by 4.2\%.
}


\maketitle

\section{Introduction}

\begin{figure*}
\centering
\includegraphics[width=0.6\textwidth]{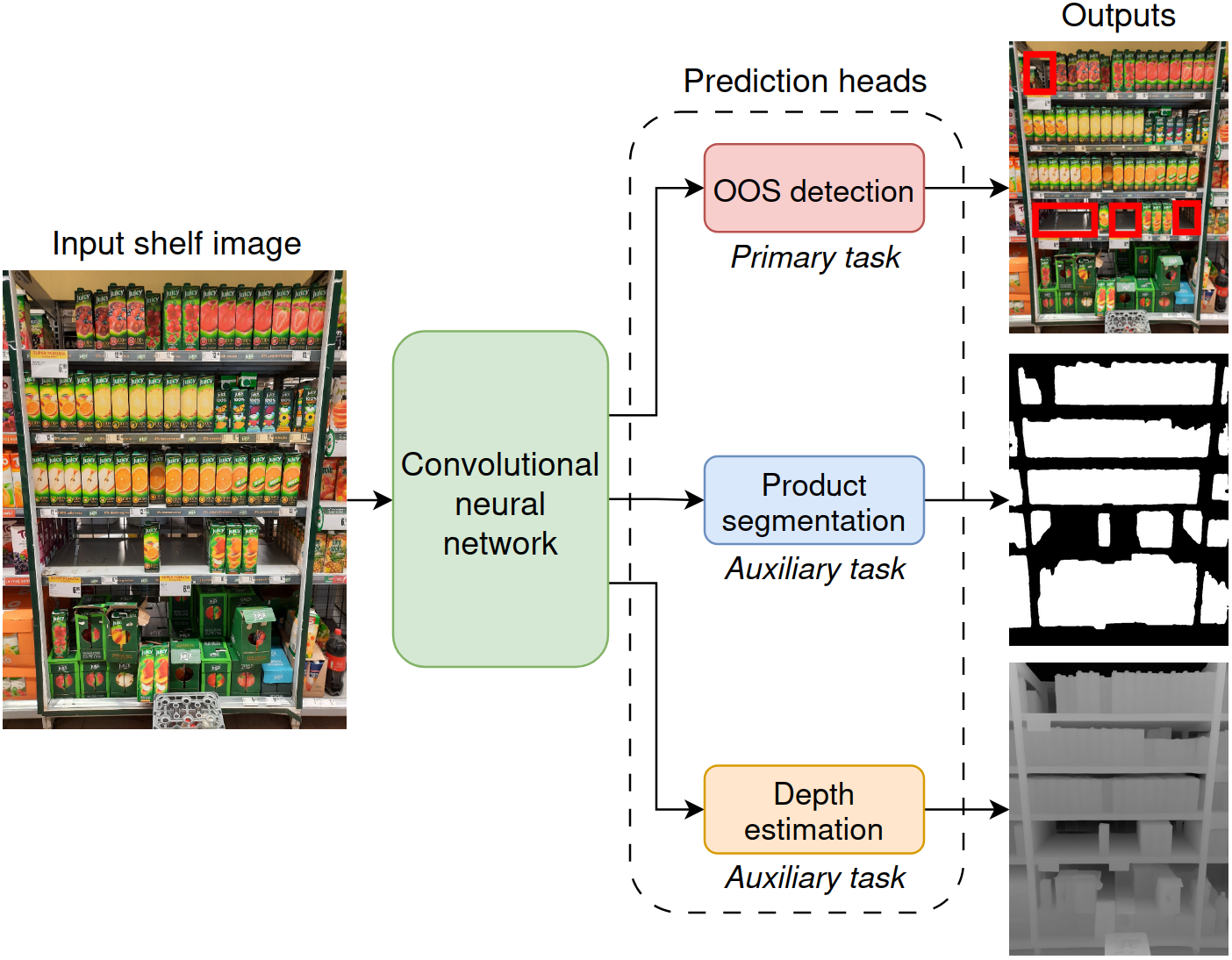}
\caption{The main idea behind OOS-DSD. The method relies on auxiliary learning to enhance OOS detection performance, with auxiliary tasks being product semantic segmentation and scene depth estimation.}
\label{fig1}
\end{figure*}

Retail stores play a crucial role in everyday shopping in modern consumeristic societies due to convenience, accessibility, and consistency. To optimize store management, supermarket chains have started to employ automated methods for various shelf analysis tasks, such as product orientation detection \cite{brane}, on-shelf availability estimation \cite{higa}, product assortment selection \cite{product_assortment}, and price tag analysis \cite{price_tag_analysis}, which are mostly based on computer vision and deep learning (DL). Furthermore, retailers also leverage out-of-stock (OOS) detection methods, whose goal is to localize a complete absence of products on the shelf, to increase sales and maintain customer satisfaction.

Although OOS detection may be performed based on stock and sales data as in \cite{buzzard, papakiri}, such methods are not completely reliable due to possible 'inventory freezing', a phenomenon in which missing products are not replenished because the detection system believes that they are not OOS \cite{inventory_freezing}.
Alternative OOS detection approaches include the use of sensors, such as radio frequency identification (RFID) \cite{rfid}. However, such approaches are expensive and suffer from low scalability. Therefore, numerous OOS detection studies focused on image analysis using machine learning (ML) \cite{rosado} or DL \cite{chen, jha, yilmazer}.

In this paper, we present OOS-DSD (\textit{\textbf{D}etect-\textbf{S}egment-\textbf{D}epth}), a novel DL-based OOS detection method that relies on the utilization of auxiliary tasks to enhance detection performance in retail shelf images. In particular, we design our convolutional neural network (CNN) architecture by extending a popular object detection model YOLOv8 \cite{yolov8} with two additional branches to achieve simultaneous OOS detection, product semantic segmentation, and scene depth estimation. The proposed model, illustrated in Fig. \ref{fig1}, is motivated by the success of multitask architectures, such as Mask R-CNN \cite{mask_rcnn}, in which the integration of an auxiliary task into the model improved the results on the primary task.
To develop OOS-DSD, we utilize our in-house retail dataset, which consists of shelf images and the corresponding annotations for each task: manually annotated OOS detection instances and product semantic segmentation maps, and pseudo-labeled relative depth maps produced by the Depth Anything V2 \cite{depth_anything_v2} depth estimation model. The integration of auxiliary learning into the method enabled OOS-DSD to surpass the performance of existing OOS detection methods.

To the best of our knowledge, our contribution is three-fold:
\begin{itemize}
    \item We propose OOS-DSD, a novel method for the simultaneous estimation of three different tasks from retail shelf images: OOS detection, product segmentation, and retail scene depth estimation. Our method achieves state-of-the-art OOS detection results on our in-house retail dataset.
    \item We show that the addition of auxiliary tasks, such as product segmentation and scene depth estimation, enhances the results on the primary task of OOS detection from shelf images.
    \item We present a depth normalization procedure for relative depth maps of retail scenes. We show that the usage of normalized data ensures better generalization of the models trained using such data. 
\end{itemize}

\section{Related work}

\subsection{Out-of-stock detection}

During the last decade, various sensors have been employed to detect OOS in supermarkets, including infrared sensors \cite{infrared}, RFID tags \cite{rfid}, etc. However, most of the proposed solutions rely on RGB cameras and different image analysis and processing techniques.
In \cite{rosado}, an ML-based method for OOS detection in shelves' panoramas was introduced. Particularly, the density of the interest points detected by the FAST corner detection \cite{fast} algorithm was used to segment the void shelf areas. The segmented areas were then filtered using color-based OOS statistics to determine OOS candidates. Finally, a support vector machine (SVM) \cite{svm} classified the valid candidates using numerous color, geometry, and texture features.
Furthermore, the authors of \cite{allegra} proposed a method that used the UNet \cite{unet} architecture to estimate OOS heatmaps from shelf images.
In \cite{santra}, a method for segmentation of void shelf areas was designed. Within the method, the input image was transformed into a superpixel graph, whose node and edge embeddings were obtained using a graph convolutional network \cite{graph_cnn} and a Siamese neural network \cite{siamese}, respectively. The extracted embeddings, along with the graph adjacency matrix, were passed to a structural SVM \cite{structural_svm} for the classification of nodes into void or non-void areas.

In recent years, OOS detection methods mostly employ object detection CNNs. In one of the first such approaches \cite{chen}, the authors performed OOS detection through product detection and analysis of areas without products. Specifically, following product detection using Faster R-CNN \cite{faster_rcnn}, image areas where products were not localized were analyzed using several independent approaches, including the presence of edges detected by the Canny operator \cite{canny}, SVM-based classification of texture features, and color histogram matching with OOS templates.
Later, the focus of object detector-based methods shifted to a direct OOS detection paradigm.
In \cite{yilmazer}, a product and OOS detection model was developed using semi-supervised learning. In particular, a small annotated dataset was utilized to train the YOLOv4 \cite{yolov4} detection model, which was later used to produce pseudo-labels for a larger unlabeled set of shelf images. The aforementioned detection model was then trained using both smaller and larger datasets.
Moreover, the EfficientDet \cite{efficientdet} and YOLOv5 \cite{yolov5} models were employed in \cite{jha} for direct OOS detection in retail images.
The authors of \cite{sikic} developed a direct OOS detection model through a two-stage training procedure, where the first stage included pre-training on augmented images, while the second stage involved fine-tuning on original data. Additionally, the aforementioned study proposed aspect ratio-based post-processing of the detected instances to remove false positives.
In \cite{robot_system}, the authors designed a framework for automatic OOS detection using an autonomous mobile robot. Within the proposed solution, YOLOv6 \cite{yolov6} was utilized to detect completely and partially empty shelf areas in an image. Furthermore, an inventory management system was proposed in \cite{managment_system}, with YOLOv8 responsible for the localization of instances and their classification into OOS and low-stock classes.

\subsection{Multitask image processing and analysis}

\begin{figure*}
\centering
\includegraphics[width=0.8\textwidth]{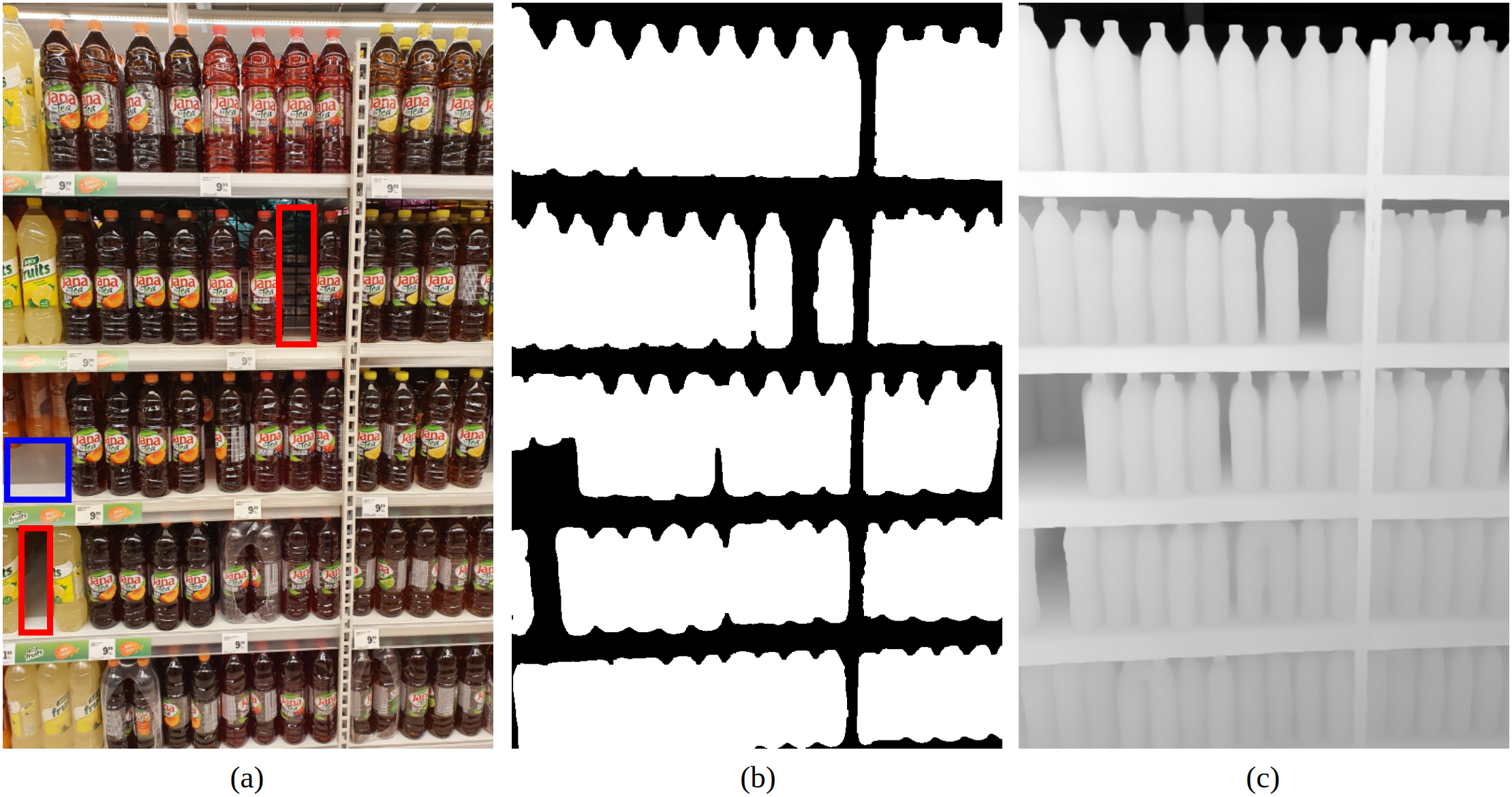}
\caption{An example of an image and annotations for each of the addressed tasks: (a) shelf image and OOS bounding boxes, (b) segmentation map, and (c) depth map. In (a), normal and front OOS classes are marked with red and blue bounding boxes, respectively.}
\label{fig_gt_maps}
\end{figure*}

Most of the neural network-based models for image processing and analysis are specialized to perform a single task, such as image classification or semantic segmentation. However, several models that perform two or more different tasks have also been proposed. In \cite{mask_rcnn}, a segmentation branch was added to the Faster R-CNN \cite{faster_rcnn} object detection model, achieving simultaneous instance detection and segmentation.
Furthermore, object detection was combined with instance depth estimation in \cite{det_dep} and with semantic segmentation in \cite{det_seg}.
Moreover, semantic segmentation and scene depth estimation maps were predicted using encoder-decoder networks in \cite{seg_dep_conv, seg_dep_trans}, with a convolutional encoder employed in \cite{seg_dep_conv} and a transformer encoder utilized in \cite{seg_dep_trans}.
Finally, several papers explored the possibility of predicting three different tasks. Particularly, \cite{zhao_three_task} presented an architecture for saliency object detection, scene depth estimation, and contour extraction, \cite{chen_three_task} addressed semantic segmentation, object detection, and instance depth estimation, whereas \cite{wang_three_task} proposed a model whose tasks were object detection, semantic segmentation, and scene depth estimation.
Although multiple multitask image analysis models have been proposed, to the best of our knowledge, there is no existing method for OOS detection in retail images that leverages the multitask training process.

\section{Dataset} \label{dataset}

Currently, there are no publicly available OOS detection image datasets with bounding box annotations necessary for training object detection models. The only dataset closely related to this task was presented in \cite{allegra}. However, in that dataset, the OOSs were annotated using a single point (i.e., there is no height and width information), making it useless for training object detection networks. Therefore, in this research, we decided to utilize our in-house OOS detection dataset, which was presented in \cite{sikic}. In particular, we conducted our research on the beverage subset of the aforementioned dataset. The subset contains 171 retail shelf images of varying resolution, which display two OOS classes: normal and front. While the normal OOS class refers to the completely void areas (from top to bottom and from front to back) on the shelf, the front OOS class indicates a scenario in which there are multiple missing products at the front of the shelf and some products still present at the back of the shelf. In total, the subset consists of 291 and 151 normal and front OOS instances, respectively. Furthermore, for this research, we extended the aforementioned data subset with segmentation and depth annotations. Specifically, for the segmentation task, each image was manually labeled to provide semantic segmentation maps that mark pixels of products present on the shelf. For the depth estimation task, we produced pseudo-labels by passing the shelf images through the Depth Anything V2 depth estimation model and taking the obtained output as depth annotations. An example of a retail shelf image and its corresponding annotations is displayed in Fig. \ref{fig_gt_maps}.

\begin{figure*}
\centering
\includegraphics[width=0.8\textwidth]{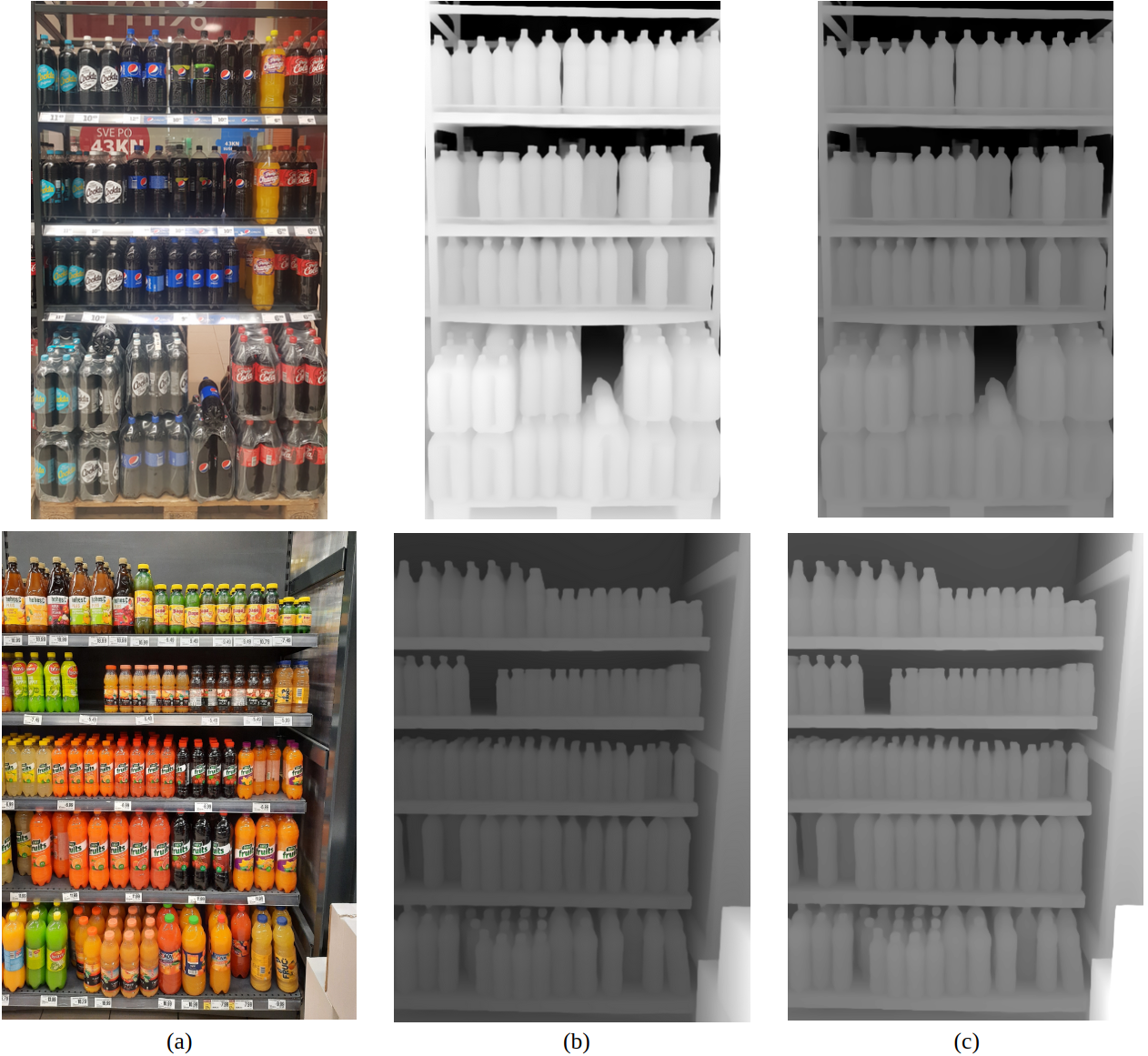}
\caption{Discrepancy between relative depth estimates of shelf images that captured scenes of different absolute depth. The columns display: (a) shelf images, (b) depth maps produced by Depth Anything V2, and (c) normalized depth maps.}
\label{fig_depth_adjusting}
\end{figure*}

\begin{algorithm*}
\caption{Depth normalization procedure}\label{alg_depth_adjust}
\begin{algorithmic}[1]
\Procedure{get\_normalized\_depth}{\textit{depth\_map\_path, segmentation\_map\_path}}
\State $\textit{depth\_map} \gets \textit{load\_image(depth\_map\_path)}$ \Comment{Grayscale image, pixels range: [0-1]}
\State $\textit{segmentation\_map} \gets \textit{load\_image(segmentation\_map\_path)}$ \Comment{Binary image: products marked with 1s, otherwise 0s}
\State $\textit{products\_depth} \gets \textit{segmentation\_map} \,\, \cdot \,\, \textit{depth\_map}$ \Comment{Element-wise multiplication}
\State $\textit{mean\_product\_depth} \gets \frac{\textit{sum(products\_depth)}}{\textit{sum(segmentation\_map)}}$ \Comment{Calculate mean depth of products}
\State $\textit{adjusted\_depth\_map} \gets \frac{\text{0.5}}{\textit{mean\_product\_depth}} \,\, \cdot \,\, \textit{depth\_map}$ \Comment{Adjust depth map to set mean depth to 0.5}
\State return \textit{adjusted\_depth\_map}
\EndProcedure
\end{algorithmic}
\end{algorithm*}

The depth maps produced by the Depth Anything V2 model contain relative depth estimates. Although our shelf images were captured from a similar distance from the shelf, the absolute depth range of the scene was sometimes substantially different between images, causing a huge discrepancy in the predicted relative depth maps. An example of this discrepancy is shown in Fig. \ref{fig_depth_adjusting}. Particularly, the top image in Fig. \ref{fig_depth_adjusting}a captured a deep background, which resulted in the shelf area being very bright in the relative depth map produced by the Depth Anything V2 model (top image in Fig. \ref{fig_depth_adjusting}b). In contrast, the bottom image in Fig. \ref{fig_depth_adjusting}a shows certain objects between the shelf and the camera, resulting in a quite dark shelf area in the relative depth map (bottom image in Fig. \ref{fig_depth_adjusting}b).

\begin{figure*}
\centering
\includegraphics[width=\textwidth]{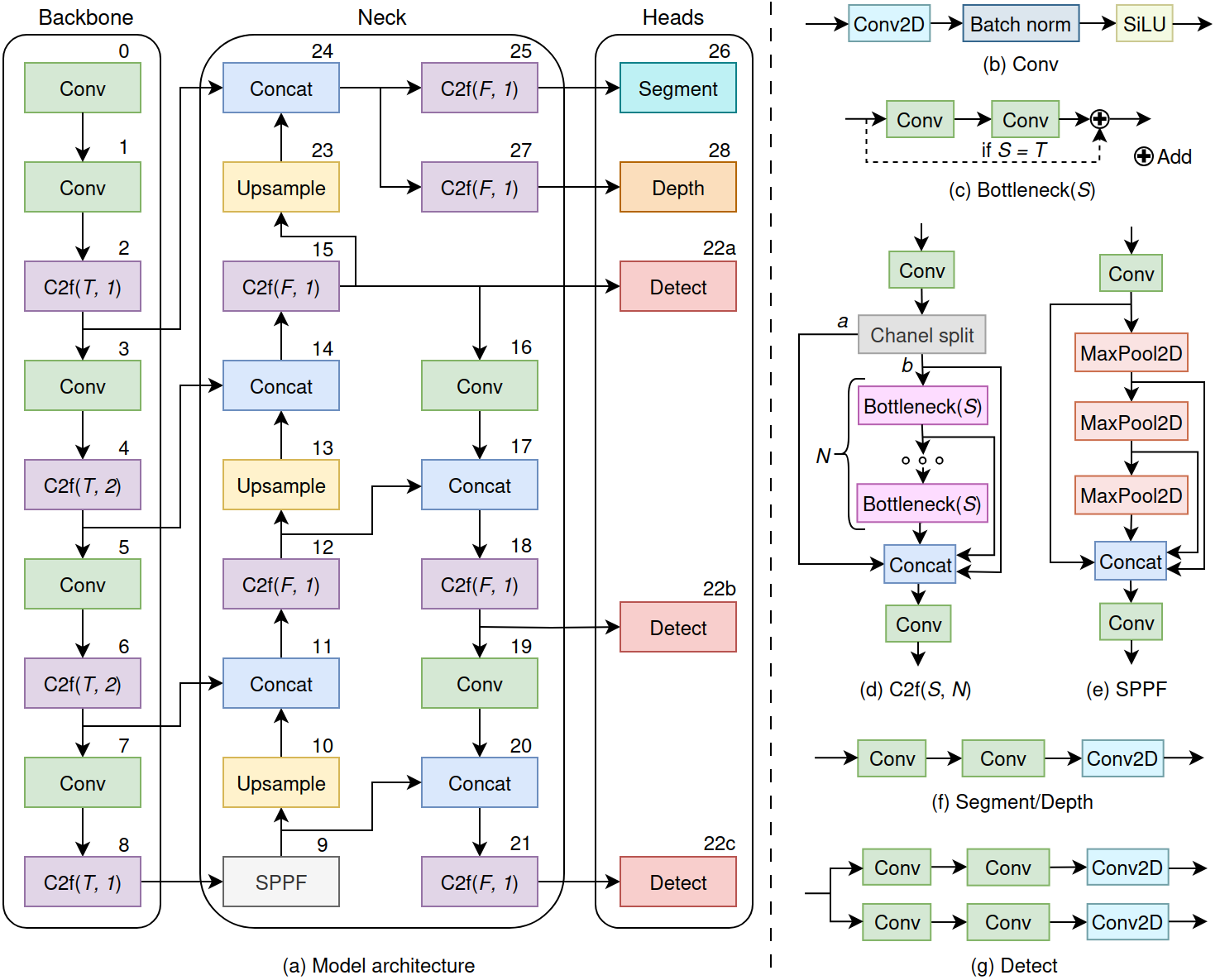}
\caption{Scheme of the proposed multitask architecture and utilized blocks: (a) overall architecture, (b) Conv block, (c) Bottleneck block, (d) C2f block, (e) SPPF block, (f) Segment/Depth heads, and (g) Detect head. In (a), each block is marked with a number shown above the block. Blocks 0-22 are the same as in the YOLOv8 model. T and F stand for true and false, respectively.}
\label{scheme}
\end{figure*}

To overcome the previously explained discrepancy, we decided to normalize the relative depth maps generated by the Depth Anything V2 model.
Specifically, the depth maps were adjusted using the pseudocode displayed in Algorithm \ref{alg_depth_adjust}. As mentioned previously, each retail shelf image is accompanied by corresponding (pseudo-)depth and segmentation maps. The input parameters to the depth normalization procedure include the file paths to these maps of a particular shelf image. The depth map image is loaded from the specified file as a grayscale image, with pixel intensities ranging from 0 (full black) to 1 (full white). Unlike the depth map image, the loaded segmentation map image is binary, i.e., contains only 1s and 0s representing product and non-product pixels, respectively. Next, the mean depth of the displayed products is calculated using the following two steps. First, the segmentation map and depth map images are multiplied element-wise, resulting in an image with 0s in the non-product pixels and [0-1] depth intensities in the product pixels. In the second step, the mean product depth is obtained by dividing the sum of the multiplication result (i.e., a cumulative depth of products) by the sum of the segmentation map image (i.e., a total number of product pixels). Finally, the depth map image is multiplied by the factor $\frac{\text{0.5}}{\text{mean product depth}}$ to adjust depth intensities in such a way that the mean product depth equals to 0.5 (i.e., central depth). Fig. \ref{fig_depth_adjusting}c shows normalized depth maps produced by applying the proposed normalization procedure to relative depth images displayed in Fig. \ref{fig_depth_adjusting}b.

\section{Methodology}

The core of the OOS-DSD method is the architecture illustrated in Fig. \ref{scheme}. The proposed architecture is designed as an extended version of the YOLOv8s, a small version of the YOLOv8 model. The architecture starts with a backbone that is comprised of alternating Conv and C2f blocks. While the Conv block represents a simple sequence of 2D convolution (Conv2D), batch normalization (Batch norm) \cite{batch_norm}, and sigmoid linear unit (SiLU) \cite{silu} activation function, the C2f block is much more complex. In particular, in Cf2, input is passed through a Conv block and split channel-wise into two halves: $a$ and $b$. Later, $b$ is refined using $N$ Bottleneck blocks, each of which consists of two sequential Conv blocks and a possible skip connection (depending on whether the parameter $S$ is set to true or false). Finally, $a$, $b$, and output from each Bottleneck block are merged by a Concat block and passed through a Conv block to form the output of the C2f block.

Next, features extracted by different backbone blocks are further bidirectionally refined in the model neck. Alongside the previously mentioned Conv, C2f, and Concat blocks, the neck contains Upsample and Spatial Pyramid Pooling Fusion (SPPF) \cite{yolov5} blocks. While the Upsample block simply increases the spatial dimension, SPPF consists of six consecutive operations. Particularly, SPPF starts with a Conv block, which is followed by a sequence of three max-pooling (MaxPool2D) layers. Output tensors from the Conv block and each of the MaxPool2D blocks are then merged by a Concat block and refined using a Conv block to produce the output of the SPPF block.

Finally, refined features from the model neck are passed through the Detect, Segment, and Depth blocks, each of which performs the prediction task specified by its name. Segment and Depth heads have the same design, which includes two consecutive Conv blocks followed by a Conv2D that produces the output segmentation/depth map. Similarly, the Detect head consists of two parallel Conv-Conv-Conv2D branches, with one branch responsible for bounding box regression and the other for instance classification.
The detection heads are applied at three different scales (following YOLOv8), while the segmentation and depth heads are applied at a single scale.

We train the proposed architecture using the dataset presented in Section \ref{dataset}. The detection output consists of bounding boxes that describe the location of recognized OOS instances. In particular, each detected instance is described as a tuple ($x, y, h, w, cls, c$) where: $x$ is the horizontal coordinate of the box's center, $y$ is the vertical coordinate of the box's center, $h$ is the height of the box, $w$ is the width of the box, $cls$ is the class of the instance, and $c$ is the confidence score of the instance. The segmentation output is a binary image with 1s at the pixels that show a product, whereas the depth estimation output is a grayscale image that shows the relative depth of each pixel.
Finally, during inference on unseen images, we follow the post-processing of \cite{sikic} to discard false positives from the detection output.

\section{Experimental results and Discussion}

\subsection{Implementation and evaluation details}

The following setup was used to train the proposed architecture. Transfer learning from the COCO \cite{coco} dataset was utilized for blocks 0-22 (i.e., the original YOLOv8 blocks) in Fig. \ref{scheme}a, while blocks 23-28 were initialized using the Kaiming uniform initialization \cite{kaiming}. We trained the model for 1000 epochs on the NVIDIA RTX 3090 GPU using SGD with Nesterov momentum \cite{nesterov} as the optimizer. The input shelf images were rescaled to 1280x1280 using letterbox scaling and augmented using horizontal flip, translation, mosaic \cite{yolov4}, and modulation of Hue-Saturation-Value channels. The augmented images were then combined into batches of 8. The total loss was calculated as the sum of the losses for each task, where the detection loss was measured as a sum of the complete-intersection-over-union loss \cite{ciou}, the distribution focal loss \cite{dfl}, and the varifocal loss \cite{vfl}, whereas the segmentation and depth estimation losses were the Dice loss \cite{dice} and the L\textsubscript{1} loss, respectively.
In the experiments, we utilized a 5-fold cross-validation, with 15\% of the training subsets left out to form validation subsets used for early stopping. When evaluating the accuracy of the models, we calculated standard metrics for each of the addressed tasks: average precision (AP)/mean AP (mAP) for class-wise/multiclass OOS detection, intersection-over-union (IoU) for product semantic segmentation, and mean absolute error (MAE) for scene depth estimation. Each quantitative result reported represents the average of the five test folds. mAP and IoU are reported in percentages.

\subsection{Comparison with state-of-the-art}

Table \ref{table_methods} provides the detection results obtained using the proposed method and a comparison to existing methods.
The existing methods are divided into two groups: methods for OOS detection via product detection and further analysis of completely void areas between the detected products (unfortunately, such a design does not allow the localization of the front OOS class instances) and methods that detect OOS directly (as OOS-DSD does).
The results show that OOS-DSD surpassed the overall performance of the existing methods by at least 1.8\% mAP. Particularly, compared to the class-wise results of previously published OOS detection methods, the proposed method achieved an improvement of at least 2.0\% and 1.6\% AP for normal and front OOS classes, respectively.

\begin{table}
\centering
\caption{Detection results of the proposed method and existing DL-based OOS-detection methods. AP and mAP metrics are used for class-wise and overall results, respectively. The methods are separated into product detection-based (top) and direct OOS detection methods (bottom). Higher is better.}
\begin{tabular}{P{3.4cm}P{0.8cm}P{0.8cm}P{0.8cm}}
 \toprule
 \multirow{2}{*}{Method} & \multicolumn{3}{c!}{Class} \\
 \cmidrule{2-4}
 & Normal & Front & All \\
 \midrule
 Chen et al. \cite{chen} (Canny) & 65.1 & - & - \\
 Chen et al. \cite{chen} (SVM) & 58.4 & - & - \\
 Chen et al. \cite{chen} (Color & \multirow{2}{*}{62.5} & \multirow{2}{*}{-} & \multirow{2}{*}{-} \\
 histogram) & & & \\
 \midrule
 Jha et al. \cite{jha} & 84.2 & 82.4 & 83.3 \\
 Šikić et al. \cite{sikic} & 87.8 & 85.4 & 86.6 \\
 \hdashline
 OOS-DSD (ours) & \textbf{89.8} & \textbf{87.0} & \textbf{88.4} \\
 \bottomrule
\end{tabular}
\label{table_methods}
\end{table}

\subsection{Ablation studies}

To analyze the impact of each auxiliary task on the results, we conducted experiments in which we eliminated the segmentation or/and depth estimation branches from the proposed architecture. Specifically, the elimination of the segmentation branch implies the removal of blocks 25 and 26 in Fig. \ref{scheme}a, while the elimination of the depth estimation branch suggests that blocks 27 and 28 were not used. Furthermore, if neither segmentation nor depth estimation branches were used, blocks 23-28 were removed, resulting in the YOLOv8s architecture. The detection, segmentation, and depth estimation results obtained within this ablation are displayed in Table \ref{table_ablation_branches}. The achieved detection results show that the addition of a single auxiliary task provided an improvement of 1.5\% to 2.7\% mAP, while the addition of both auxiliary tasks increased the results by 3.7\% mAP. Furthermore, the incorporation of a depth estimation branch into a model that performs detection and segmentation tasks not only increased the detection results by 1.0\% mAP but also had a positive impact on the segmentation results, which were improved by 0.7\% IoU. Similarly, the extension of the detection and depth estimation model with a segmentation branch, alongside increasing the detection results by 2.2\% mAP, improved depth estimation by reducing MAE by 0.028.

\begin{table}
\centering
\caption{Ablation of the branch/es used in the architecture. Det, Seg, and Dep stand for detection, segmentation, and depth, respectively. $\uparrow$ and $\downarrow$ denote that higher and lower is better, respectively.}
\begin{tabular}{P{0.7cm}P{0.7cm}P{0.7cm}P{0.7cm}P{0.7cm}P{0.7cm}}
 \toprule
 \multicolumn{3}{c!}{Branch used} & \multirow{2}{*}{mAP $\uparrow$} & \multirow{2}{*}{IoU $\uparrow$} & \multirow{2}{*}{MAE $\downarrow$} \\
 \cmidrule{1-3}
 Det & Seg & Dep & & & \\
 \midrule
 \cmark & & & 84.7 & - & - \\
 \cmark & \cmark & & 87.4 & 91.7 & - \\
 \cmark & & \cmark & 86.2 & - & 0.129 \\
 \cmark & \cmark & \cmark & 88.4 & 92.4 & 0.101 \\
 \bottomrule
\end{tabular}
\label{table_ablation_branches}
\end{table}

Since different loss functions may be used to supervise both the segmentation and depth estimation parts of the output, it is necessary to examine the influence of various losses on the performance when developing a model.
Table \ref{table_ablation_losses} displays an ablation for the multiple segmentation and depth estimation losses evaluated. In particular, we have tested binary cross-entropy (BCE), mean squared error (MSE), L\textsubscript{1}, and Dice losses for the segmentation task, whereas MSE and L\textsubscript{1} losses were evaluated for the depth estimation task. The obtained results demonstrate the superiority of Dice loss in the segmentation task and L\textsubscript{1} loss in the depth estimation task. Particularly, when the Dice segmentation loss was coupled with either the MSE or L\textsubscript{1} loss for depth estimation, an improvement of at least 0.7\% mAP and 0.4\% IoU was achieved compared to the alternatives. Moreover, when the L\textsubscript{1} loss was utilized for depth estimation and coupled with any segmentation loss, an increase of at least 1.2\% mAP and a reduction of at least 0.009 MAE was obtained compared to using the MSE loss.

\begin{table}
\centering
\caption{Ablation of the losses used to supervise the segmentation (Seg) and depth estimation (Dep) branches. $\uparrow$ and $\downarrow$ denote that higher and lower is better, respectively.}
\begin{tabular}{P{0.8cm}P{0.8cm}P{0.7cm}P{0.7cm}P{0.7cm}}
 \toprule
 \multicolumn{2}{c!}{Branch loss} & \multirow{2}{*}{mAP $\uparrow$} & \multirow{2}{*}{IoU $\uparrow$} & \multirow{2}{*}{MAE $\downarrow$} \\
 \cmidrule{1-2}
 Seg & Dep & & & \\
 \midrule
 BCE & MSE & 86.4 & 91.3 & 0.113 \\
 MSE & MSE & 86.0 & 90.7 & 0.116 \\
 L\textsubscript{1} & MSE & 85.5 & 90.1 & 0.116 \\
 Dice & MSE & 87.2 & 91.9 & 0.112 \\
 BCE & L\textsubscript{1} & 87.7 & 92.0 & 0.103 \\
 MSE & L\textsubscript{1} & 87.5 & 91.6 & 0.104 \\
 L\textsubscript{1} & L\textsubscript{1} & 87.3 & 91.5 & 0.107 \\
 Dice & L\textsubscript{1} & 88.4 & 92.4 & 0.101 \\
 \bottomrule
\end{tabular}
\label{table_ablation_losses}
\end{table}

Finally, we inspect the effect of the depth normalization procedure (described in Section \ref{dataset}) on the results. In particular, we have evaluated two different models capable of depth prediction: the proposed OOS-DSD model and a version of the OOS-DSD that does not have the segmentation branch (OOS-DSD w/o Seg.). The aforementioned models were trained using data that included either original or normalized depth maps. The obtained evaluation results are displayed in Table \ref{table_ablation_depth}. In this ablation, the MAE depth estimation results were not calculated since the depth normalization procedure needed to be performed across the entire dataset, i.e., it changed the testing depth maps as well, thus disabling the possibility of a valid depth estimation comparison between models. The results reveal a strong impact of the presented depth normalization procedure, which enabled a 4.6\% increase in mAP for the model without a segmentation branch, whereas the performance of OOS-DSD was improved by a 4.2\% mAP and 3.5\% IoU.

\begin{table}
\centering
\caption{Ablation on applying the depth normalization procedure to depth maps. Higher is better.}
\begin{tabular}{P{2.6cm}P{1.8cm}P{0.7cm}P{0.7cm}}
 \toprule
 \multirow{2}{*}{Model} & Depth & \multirow{2}{*}{mAP} & \multirow{2}{*}{IoU} \\
 & normalization & & \\
 \midrule
 OOS-DSD w/o Seg. & \xmark & 81.6 & - \\
 OOS-DSD & \xmark & 84.2 & 88.9 \\
 OOS-DSD w/o Seg. & \cmark & 86.2 & - \\
 OOS-DSD & \cmark & 88.4 & 92.4 \\
 \bottomrule
\end{tabular}
\label{table_ablation_depth}
\end{table}

\subsection{Qualitative results}

\begin{figure*}
\centering
\includegraphics[width=0.88\textwidth]{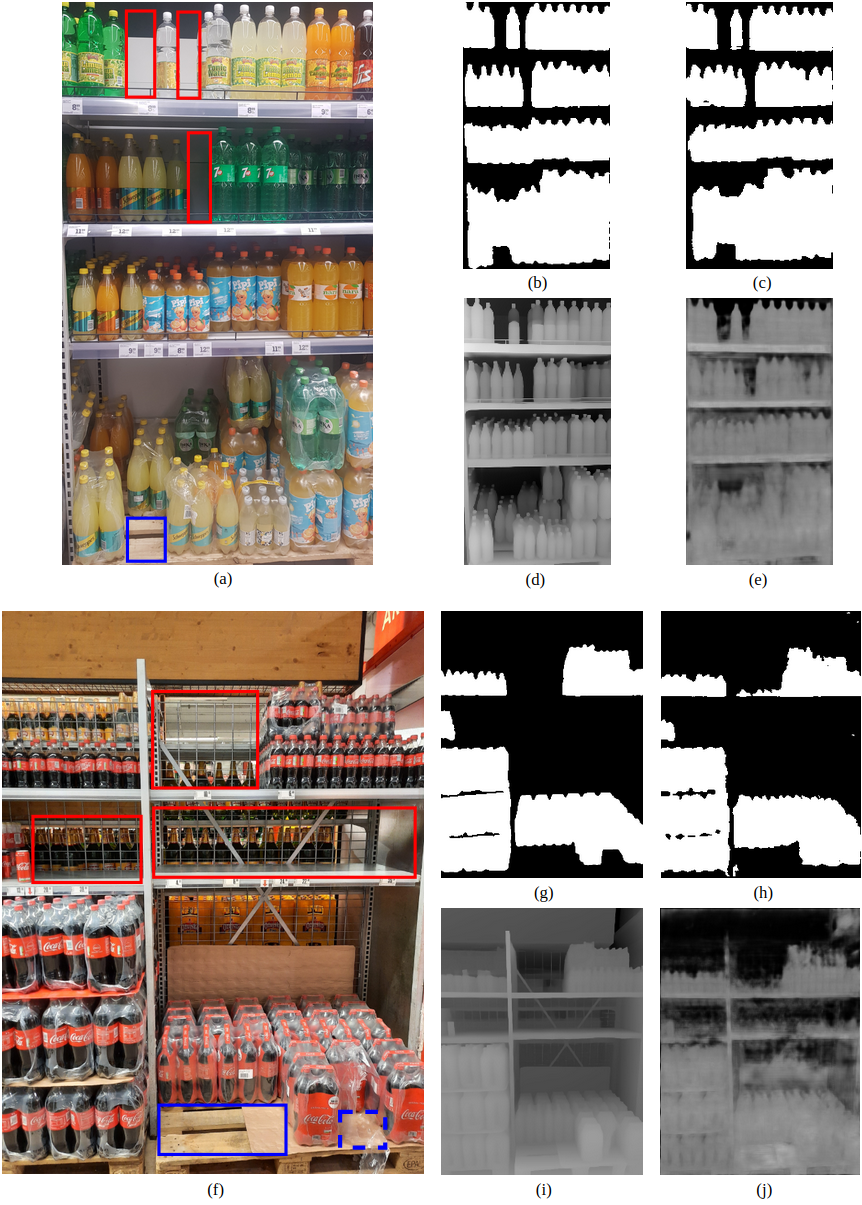}
\caption{Examples of the ground truth data and predictions by the OOS-DSD for two shelf images: (a, f) OOS detection results, (b, g) ground truth segmentation maps, (c, h) predicted segmentation maps, (d, i) normalized (pseudo-ground truth) depth maps, and (e, j) predicted depth maps. Output results of the two images are grouped (a-e) and (f-j). In (a, f), red color marks the normal OOS class, blue color marks the front OOS class, full line denotes accurate detection, and dashed line denotes not recognized OOS. }
\label{fig_qualitative}
\end{figure*}

Fig. \ref{fig_qualitative}a-e displays an example of a very good model output. The detection result was completely accurate as the model correctly localized each OOS instance (three normal OOS instances and one front OOS instance). Furthermore, the predicted binary segmentation map had an excellent overlap with the ground truth segmentation map, with only a few very small non-product blobs being inaccurately predicted within the product areas. Finally, in the depth estimation task, the model is generally able to distinct the depth of the shelf, products, and background, although the prediction is quite noisy compared to the ground truth. However, such a performance could be expected when training on pseudo-labeled data. 

Although the model generally performs very well, there were some shelf images for which the model was unable to produce an accurate output. An example of such an image is shown in Fig. \ref{fig_qualitative}f-j, with certain problems in each prediction task. Particularly, both detection and segmentation tasks failed in the bottom right corner of the image, where the model hallucinated the existence of products. Furthermore, the segmentation head incorrectly predicted the presence of some products on the top of the shelf, in addition to several product blobs in areas without products across the whole image. Finally, the depth estimation head struggled with products visible through the shelf, estimating their depth as if they were positioned on the captured shelf rather than behind it.

\section{Conclusion}

In this paper, we have presented OOS-DSD, a DL-based method that enhances OOS detection in retail images through the use of auxiliary learning.
The proposed method achieved excellent OOS detection results, outperforming existing methods by at least 2.0\% and 1.6\% higher AP for normal and front OOS classes, respectively.
Auxiliary learning, i.e., the incorporation of product segmentation and scene depth estimation tasks into the method, ensured a tremendous positive impact on OOS detection performance, increasing mAP by 3.7\%.
Furthermore, the presented depth normalization procedure proved to be crucial for the stabilization of the training process, as it improved the OOS detection results by 4.2\% mAP.
Finally, a thorough evaluation of the loss functions revealed that a combination of the Dice segmentation loss and the L\textsubscript{1} depth estimation loss performed the best, achieving mAP up to 2.9\% higher than alternatives.
In future work, the proposed methodology could be extended to multiview settings, i.e., using multiple images of the same shelf from different viewing positions. Such an approach would allow for a much better scene perception, leading to accurate recognition of the most challenging OOS instances.

\section*{Statements and Declarations}

\textbf{Competing Interests.} The authors declare that there is no conflict of interest.

\bibliography{sn-bibliography}

\end{document}